\RequirePackage{fix-cm}
\documentclass[smallextended]{svjour3}       % onecolumn (second format)
\smartqed  % flush right qed marks, e.g. at end of proof
\usepackage{graphicx}
\usepackage{times}
\usepackage{epsfig}
\usepackage{amsmath}
\usepackage{amssymb}
\usepackage{bm}
\usepackage{soul}
\usepackage{xcolor}
\usepackage{caption}
\usepackage{subcaption}
\usepackage{booktabs, multirow}
\usepackage{array}
\usepackage[export]{adjustbox}
\usepackage{hyperref}
\usepackage[T1]{fontenc}
\usepackage{subcaption}
\usepackage{bbm}
\newcolumntype{P}[1]{>{\centering\arraybackslash}p{#1}}
\newcommand{\normtwo}[1]{\left\lVert#1\right\rVert_2}
\newcommand{\normone}[1]{\left\lVert#1\right\rVert_1}
\newcommand{\etal}{\emph{et al.}{} }%

\renewcommand{\epsilon}{\varepsilon}
\newcommand{\gradient}{\| \mathcal{J}_{c}(\bm x) \|_F}
\newcommand{\mat}[1]{\bm{\mathrm{#1}}}

\journalname{SN Computer Science}
\begin{document}

\title{An Empirical Study on the Relation between Network Interpretability and Adversarial Robustness\thanks{\textit{Conflicts of interest}: The work done by Adam Noack was funded by the NSF Center for Big Learning (CBL) and a grant from the Air Force Research Laboratory and Defense Advanced Research Projects Agency, under agreement number FA8750-16-C-0166, subcontract K001892-00-S05. Isaac Ahern's work was funded by the NSF CBL. Dejing Dou was originally funded by the NSF CBL and now works at Baidu. Boyang Li originally worked with Baidu, but now works at Nanyang Technological University.
}} 
% Grants or other notes
% about the article that should go on the front page should be
% placed here. General acknowledgments should be placed at the end of the article.}
% }
% \subtitle{}

\titlerunning{Relating Network Interpretability and Adversarial Robustness}        % if too long for running head

\author{Adam Noack \and Isaac Ahern \and Dejing Dou \and Boyang Li$^*$
}

% \authorrunning{Short form of author list} % if too long for running head

\institute{Adam Noack  \at
              University of Oregon \\
              \email{anoack2@uoregon.edu}           %  \\
           \and
           Isaac Ahern \at
              University of Oregon \\ 
               \email{iahern@uoregon.edu}
           \and
           Dejing Dou \at
              University of Oregon \\ 
               \email{dou@cs.uoregon.edu}
           \and
           Boyang Li ($^*$ corresponding author)\at
             School of Computer Science and Engineering \\
             and Joint NTU-UBC LILY Research Center, \\
					Nanyang Technological University \\ 
               \email{boyang.li@ntu.edu.sg}
}

\date{Received: Jul 9, 2020 / Accepted: Oct 31, 2020}

\maketitle

\begin{abstract}
Deep neural networks (DNNs) have had many successes, but they suffer from two major issues: (1) a vulnerability to adversarial examples and (2) a tendency to elude human interpretation. Interestingly, recent empirical and theoretical evidence suggests these two seemingly disparate issues are actually connected. In particular, robust models tend to provide more interpretable gradients than non-robust models. However, whether this relationship works in the opposite direction remains obscure.
With this paper, we seek empirical answers to the following question: can models acquire adversarial robustness when they are trained to have interpretable gradients? 
We introduce a theoretically inspired technique called \emph{Interpretation Regularization} (IR), which encourages a model's gradients to (1) match the direction of interpretable target salience maps and (2) have small magnitude. 
To assess model performance and tease apart factors that contribute to adversarial robustness, we conduct extensive experiments on MNIST and CIFAR-10 with both $\ell_2$ and $\ell_\infty$ attacks. We demonstrate that training the networks to have interpretable gradients improves their  robustness to adversarial perturbations. Applying the network interpretation technique SmoothGrad \cite{smoothgrad} yields additional performance gains, especially in cross-norm attacks and under heavy perturbations. The results indicate that the interpretability of the model gradients is a crucial factor for adversarial robustness. Code for the experiments can be found at \url{https://github.com/a1noack/interp\_regularization}.

% Insert your abstract here. Include keywords, PACS and mathematical
% subject classification numbers as needed.

\vspace{.15cm}
\noindent \textbf{Declarations}

\textbf{Funding} This work was funded by the NSF Center for Big Learning and a grant from the Air Force Research Laboratory and Defense Advanced Research Projects Agency, under agreement number FA8750-16-C-0166, subcontract K001892-00-S05.

\textbf{Availability of data and material} All data and materials we used for our experiments are freely available via PyTorch's torchvision package \cite{pytorch}.

\textbf{Code availability} All of the code used for our experiments can be found at this url: 
\url{https://github.com/a1noack/interp\_regularization}.

\vspace{.15cm}
\noindent \textbf{Acknowledgements} Funding for this project was provided by the National Science Foundation Center for Big Learning and by the Defense Advanced Research Projects Agency's Media Forensics grant.

\end{abstract}

\section{Introduction} \label{sec:intro}
Over the past decade, deep neural networks (DNNs) have produced unprecedented results across a wide range of tasks. However, their impressive performance has been clouded by two weaknesses: (1) their susceptibility to adversarial perturbations and (2) their opaque decision making process.
% difficulty in interpreting how they reach their decisions. 
These weaknesses erode users' trust in DNNs and limit DNN adoption in security-critical applications, yet our understanding of these two weaknesses is still limited. With this paper, we explore the potential connection between the two phenomena.  

Adversarial perturbations are small, almost imperceptible changes to an input that cause a machine learning model to make erroneous predictions \cite{intriguing_properties_dnns}. Many attacks that can efficiently find such perturbations have been developed recently, including the fast gradient sign method (FGSM) \cite{explaining_harnessing_adv_egs}, projected gradient descent (PGD) \cite{toward_robust_dnn}, the Carlini-Wagner attack \cite{cw}, and many others \cite{blackbox_attack,Brendel2017:Decision-based-attacks,universal-perturbations2017,Chen2017:zoo}.
In response, many defense techniques have been proposed \cite{explaining_harnessing_adv_egs,distillation,input_grad_reg,double_bp,lipschitz_converge_generalize,jacobian_reg,smoothing}. Despite the large volume of published work in this area, to date the best defenses remain imperfect, and the cause for the existence of adversarial perturbations continues to be a debated topic \cite{Gilmer2018,Schmidt2018,Bubeck2018,Nakkiran2019,ilyas2019adversarial}.

A second weakness of DNNs is their opaqueness; even human experts struggle to explain the underlying rationales for DNNs' decisions. The black-box nature of DNNs is especially undesirable in domains such as medicine and law where the reasoning process used to arrive at a decision is often just as important as the decision itself. This need for DNN interpretability has led to the development of interpretation techniques that identify features used by a network to make its predictions \cite{smoothgrad,deeplift,ahern2019normlime}, to visualize the network weights \cite{Zeiler2014,BoleiZhou2015,NetDissection2017,fong2018}, or to calculate the influence of training data influence on its decisions \cite{Koh2017:influence-functions}. These techniques contribute to the unmasking of the complex mechanisms that underlie DNN behaviors, but by and large DNNs remain incomprehensible black boxes.

Adversarial vulnerability and model opaqueness were previously assumed to be unrelated. However, recent results suggest that the two issues may be connected; specifically, several works have demonstrated, mostly qualitatively, that robust DNNs tend to be interpretable. Tsipras \etal \cite{robustness_vs_accuracy} found that the loss gradient with respect to the input of adversarially trained networks visually align with the features that humans intuitively understand to be salient. Similarly, it has been noticed that gradient regularization \cite{input_grad_reg} and Lipschitz constraints \cite{lipschitz_nets_interp}, both of which improve adversarial robustness, lead to qualitatively interpretable gradient maps.
Etmann \etal \cite{robustness_interpretability} theoretically showed that, for linear models, Lipschitz regularization causes the gradients to align with the input images. These results constitute a converging collection of evidence that optimizing a network for robustness leads to some degree of interpretability.

With this paper, we explore the other direction of the causality and seek answers to the converse question: \emph{if a network is trained to have interpretable gradients, will it be robust against adversarial attacks?} In the following, we offer some theoretically motivated justification for an affirmative answer. At a high level, in order to achieve good adversarial performance, we must maintain high predictive accuracy and curtail the performance degradation caused by adversarial samples at the same time. Minimizing degradation requires the norm of the Jacobian matrix to be small, whereas high generation performance requires the Jacobian to capture data regularities. We postulate that an interpretable Jacobian may strike the right balance. 

% We give a theoretically motivated explanation why an interpretable Jacobian can be conducive to good adversarial performance. 

We first consider the minimization of performance degradation caused by adversarial samples and its implications on the network Jacobian. 
For a given input $\bm x_0 \in \mathbb{R}^D$ and its one-hot encoded label $\bm y_0 \in \mathbb{R}^K$, we adopt a neural network $f(\cdot)$ with ReLU activation, from which the final prediction is $\hat{\bm y}_0 = F(\bm x_0) =\text{softmax}(f(\bm x_0))$. 
% Using $\mat{W}^l$ to denote the parameter matrix at $l^{\text{th}}$ layer, the network operation can be written as 
% \begin{equation}
% f(\bm x) = \mat{W}^L\, \text{ReLU}(\mat{W}^{L-1}\, \cdots \,\text{ReLU}(\mat{W}^1 \bm x_0))
% \end{equation}
% It is worth noting that the function $f(\bm x)$ in the neighborhood of $\bm x$ can be written as a sequence of matrix multiplications and is completely linear because ReLU can be understood as zeroing out matrix rows depending on $\bm x$.
During an $\ell_2$ adversarial attack, the adversary seeks a small perturbation $\bm{\delta}$ to the input $\bm{x}_0$ such that the prediction will change significantly and $\normtwo{\bm{\delta}}$ is smaller than a predefined threshold. 
Let $\rho(\bm{x})$ be a lower bound on the change in the confidence norm necessary to flip the prediction of $F(\bm{x})$. For the attack to be successful, we must have 
$\normtwo{F(\bm{x}_0 + \bm{\delta}) - F(\bm{x}_0)} > \rho(\bm{x}_0)$. Letting the Jacobian of the whole network be denoted by $\mat{J}(\bm x_0) = \partial \hat{\bm y}_0 / \partial \bm x_0$, we can apply first-order Taylor expansion and yield 
\begin{equation}
   \normtwo{\mat{J}(\bm x_0) \bm{\delta}} > \rho(\bm{x}_0)
\end{equation}
Under the singular value decomposition, $\mat{J}(\bm x_0) = \mat{U}\mat{\Sigma}\mat{V}^\top$. We let $\bm s$ be the vector of singular values on the diagonal such that $\mat{\Sigma} = \text{diag}(\bm{s})$ and derive\footnote{Here we assume $K=D$ for simplicity. The common case $K<D$ is very similar but involves the slightly more complex notation for truncating $\mat{V}^\top \bm{\delta}$.} 
\begin{equation}
   \normtwo{\mat{J}(\bm x_0) \bm{\delta}} = \normtwo{ \bm s \otimes \mat{V}^\top \bm{\delta} } 
\end{equation}
where $\otimes$ denotes the Hadamard product. With slight abuse of notation, let $|a|$ denote the absolute value of scalar $a$ and $|\bm u|$ denote the element-wise absolute value operation for vector $\bm u$. By the relation between $\ell_2$ and $\ell_1$ norms,
\begin{equation}
   |\bm{s}|^\top |\mat{V}^\top \bm{\delta}| = \normone{ \bm s \otimes \mat{V}^\top \bm{\delta} } \ge \normtwo{ \bm s \otimes \mat{V}^\top \bm{\delta} } 
\end{equation}
Therefore, to make sure the attack is unsuccessful, it is sufficient to have
\begin{equation}
   \rho(\bm{x}_0) \ge |\bm{s}|^\top |\mat{V}^\top \bm{\delta}|= \normtwo{\bm s}  \normtwo{ \bm{\delta}} \cos(|\bm s|, |\mat{V}^\top\bm{\delta}|) 
\end{equation}
\sloppypar{\noindent 
where $\cos(\cdot, \cdot)$ is the cosine of the angle between the two vectors. Adversarial attack is a Stackelberg game \cite{StacklebergGame:2011} where the attacker chooses the perturbation $\bm{\delta}$ after the network is trained. Hence, for good defense we need to minimize $\normtwo{\bm s}$ and the worst-case cosine $\max_{\bm{\delta}} \cos(|\bm s|, |\mat{V}^\top\bm{\delta}|)$.
Jacobian regularization minimizes $\|\mat{J}(\bm x_0)\|_F$, which is equivalent to $\| \bm s \|_2$. However, given a full rank $\mat{V}$, the adversary can always find a perturbation $\bm{\delta}$ that maximizes $\cos(|\bm s|, |\mat{V}^\top\bm{\delta}|)$, so it may seem that selecting the direction of $\bm s$ is futile. }

Nevertheless, good adversarial performance requires more than just minimizing the damage caused by adversarial examples $\normtwo{F(\bm{x}_0 + \bm{\delta}) - F(\bm{x}_0)}$. Adversarial performance may still be poor if the network is equally mistaken on normal and adversarial examples. 
%When $\bm s = c\mathbbm{1}$ is uniform with constant $c$, the Jacobian $\mat{J}(\bm x_0) = c\mat{U}\mat{V}^\top$ degenerates to a simple rotation matrix with constant scaling, unlikely to extract useful features from $\bm x_0$. 
In fact, some analyses of the learning dynamics of neural networks \cite{Saxe2019,lampinen2019analytic} emphasize the role of the Jacobian singular values in generalization. The network first learns large singular values before small singular values, which may be affected by noise. This indicates that, to maximize generalization performance, the direction of $\bm s$ must not be arbitrary and not uniform. 

If $\bm s$ is not uniform, then it necessarily amplifies certain dimensions of $\mat{V}^\top \bm x_0$ while attenuating others, which is a form of feature selection.
Ilyas \etal \cite{ilyas2019adversarial} have noticed the issue of feature selection and argue that DNNs are brittle because they use non-robust features, which are correlated to the class label but incomprehensible for humans. Intuitively, selectively spending the limited budget of $\| \bm s \|_2$  on features that are invariant to adversarial perturbations and comprehensible to humans should improve adversarial accuracy. Therefore, we offer the conjecture that interpretable gradients strike the right balance between high generalization performance and low adversarial degradation. 

%Jacobian regularization forces the model to constrain the weights it learns so that the gradients $\| \frac{\partial f(\bm{x})}{\partial \bm{x}} \| |_{\bm{x} = \bm{x}_0}$ are small, smoothing out the local geometry of the loss function and pushing the decision boundary away from points that are candidate adversarial examples. 

We propose to train models to match interpretable gradients, which we call Interpretation Regularization (IR). In order to obtain interpretable gradients, we extract gradient-based interpretations from adversarially trained robust models, which provide more human-like interpretations than non-robust models, and use them as targets during training. We demonstrate that IR improves model robustness and outperforms Jacobian regularization, despite the fact that our method only acts on one column of the input-output Jacobian rather than the entire Jacobian matrix \cite{jacobian_reg,jacobian_reg_improves_generalization,jacobian_reg_approx}. Most importantly, target interpretations extracted by SmoothGrad \cite{smoothgrad}, which are smoother and more interpretable than simple gradients, lead to further robustness gains, especially in difficult cases like cross-norm attacks and large perturbations. This indicates that Interpretation Regularization is more than just distilling existing robust models. 

% \hl{
% Through further analysis, we discover that two factors contribute to the effectiveness of Interpretation Regularization (IR): the suppression of the gradient magnitude and the selective use of features guided by high-quality interpretations. The two factors can explain model behaviors under various settings of regularization and target interpretation. }

It is worth emphasizing that the paper's contribution is in highlighting the connection between interpretability and adversarial robustness. Interpretation Regularization does not and is not intended to provide a practical adversarial defense because it requires an adversarially trained robust model to supply a target interpretation. More specifically, our contributions are:
\begin{itemize}
    \itemsep0em
    \item We empirically investigate if networks optimized to have interpretable gradients are robust to adversarial attacks. We find that simply requiring the model to match interpretations extracted from a robust model can improve robustness. Applying the network interpretation technique SmoothGrad further reinforces robustness. 
    \item To explain the experimental results, we analyze the connection between Jacobian regularization and Interpretation Regularization. We identify two factors---the suppression of the gradient and the selective use of features guided by high-quality interpretations---that contribute to the effectiveness of Interpretation Regularization and explain model behaviors. 
    
\end{itemize}

\section{Related Work} \label{sec:related_work}
In this section, we provide a brief review of the vast literature on DNN interpretation, adversarial attacks, and adversarial defenses. 

\subsection{Interpretation of DNNs} \label{subsec:interpreting_dnn}
Numerous methods have been proposed to interpret and understand different aspects of DNNs. For example, the representations learned in the network layers can be probed and visualized \cite{Zeiler2014,BoleiZhou2015,NetDissection2017,fong2018}. The influence of training data on the model's prediction can be estimated \cite{Koh2017:influence-functions}. In this paper, we are mostly concerned with interpretations in the form of features' contribution to the model's prediction. When the input is an image, the measure of feature contribution is often referred to as an \emph{importance map} or a \emph{salience map}. 

The gradient of the network output with respect to the input provides a simple yet effective method for generating salience maps \cite{simple_gradient}.  The following Taylor expansion approximates the model behavior $f(\bm{x})$ around $\bm{x}$.  \begin{equation}
f(\bm x + \bm \epsilon) = f(\bm x) + \frac{\partial f(\bm{x})}{\partial \bm{x}}^\top \bm \epsilon + o(\bm \epsilon ^\top \bm \epsilon)
\end{equation}
where $\bm \epsilon$ represents a small change to $\bm x$. 
The relative importance of the feature $x_i$ can then be captured by the absolute value $|\frac{\partial f(\bm{x})}{\partial x_i} |$, which measures how $f(\bm{x})$ changes when a small change is applied to $x_i$. While such interpretations highlight salient features of an image, the simple gradient often exhibits a large degree of visual noise and does not always correspond to human intuition regarding feature contribution. %\cite{smoothgrad}. 
This has motivated the development of more elaborate salience map generation techniques in order to induce more structured and visually meaningful interpretations. These include Gradient $\times$ Input \cite{deeplift}, Integrated Gradients \cite{integrated_gradients}, Deep Taylor Decomposition \cite{deep_taylor}, DeepLIFT \cite{deeplift}, Guided Backprop \cite{guided_backprop}, and GradCAM / Guided GradCAM \cite{selvaraju2016gradcam}. SmoothGrad \cite{smoothgrad} and VarGrad \cite{sanity_checks_salience_maps} compute Monte Carlo expectations of the first and second moments of the gradient when noise is added to the input image. Contrastive explanations \cite{dhur2018explanations} identify how absent components contribute to the prediction.

Evaluation of the generated salience maps is an important and challenging topic. \cite{kindermans2017learning} analyzes behaviors of interpretation methods acting on simple linear models. \cite{sanity_checks_salience_maps} proposes that salience map generating methods should satisfy certain desiderata, including sensitivity towards model and label perturbations. \cite{unreliability} argues that salience map generating methods should be invariant to uniform mean shifts of the input. Interestingly, several popular methods (Integrated Gradients, Guided Backprop, Guided GradCAM, etc) do not satisfy these apparently reasonable requirements.

\subsection{Adversarial Attacks} \label{subsec:adv_attack_defense}

To perform an adversarial attack to a neural network $f(\bm x)$ that correctly classifies its input $\bm x$, we seek a perturbation $\bm \delta$, whose norm is less than a predefined threshold $\epsilon$, such that $f(\bm x + \bm \delta)$ outputs an incorrect classification. 
To precisely understand the nature of adversarial attacks, we create the threat model \cite{carlini2019evaluating}, which describes the attacker's goals, knowledge, and capabilities. In terms of goals, untargeted attacks do not care about the model's exact predictions as long as they are incorrect, whereas targeted attacks aim to force a particular erroneous prediction. In terms of knowledge, white-box attacks have access to the model's loss gradients, whereas black-box attacks do not. The attacker's capabilities may be modeled as the amount of perturbation they are allowed to make, usually measured using the $\ell_0$, $\ell_2$, or $\ell_{\infty}$ metric.

Among white-box attacks, the fast gradient sign method (FGSM) \cite{explaining_harnessing_adv_egs} provided a proof-of-concept by adding an $\epsilon$-scaled sign vector of the loss gradient to the input image. More formally, the untargeted FGSM updates $\bm x$ in the direction of increasing the network loss $\mathcal{L}(\bm x, \bm y, \bm \theta)$,
\begin{equation}
\bm x^\prime = \bm x + \epsilon \, \text{sign}\left(\frac{\partial \mathcal{L}(\bm x, \bm y, \bm \theta)}{\partial \bm x}\right).
\end{equation}
where the $\text{sign}(\cdot)$ function maps a $D$-dimensional vector to $\{-1, 1\}^D$ depending on the signs of its components. 
In comparison, the targeted FGSM moves $\bm x$ in the direction of decreasing the loss $\mathcal{L}(\bm x, \bm y^\prime, \bm \theta)$ for an incorrect label $\bm y^\prime$
\begin{equation}
\bm x^\prime = \bm x - \epsilon \, \text{sign}\left(\frac{\partial \mathcal{L}(\bm x, \bm y^\prime, \bm \theta)}{\partial \bm x}\right).
\end{equation}
Extending FGSM, projected gradient descent (PGD) \cite{toward_robust_dnn} provides a more powerful iterated optimization approach. Whenever the perturbation magnitude exceeds the attacker's budget, the perturbed input is projected back to the allowed range. 
The Jacobian-based Saliency Map Attack (JSMA) \cite{jsma} modifies pixels that have large gradients. DeepFool \cite{deepfool} applies a local linear approximation in iterated optimization. Carlini and Wagner \cite{cw} used constrained optimization and reparameterization to effectively search for adversarial samples. 

Effective attacks can be built even when gradient information is not available. \cite{blackbox_attack} builds a dataset by querying the target model and use the dataset to train a substitute network from which gradient can be obtained for the attack. Carefully constructed adversarial examples can be transferred across models \cite{Liu2017:transferable,xie2018improving} and across images \cite{universal-perturbations2017}. In addition, gradient-free attacks \cite{Chen2017:zoo,Uesato2018:adversarial-risk,Ilyas2018:black-box} do not rely on gradient information. Brendel \etal \cite{Brendel2017:Decision-based-attacks} proposed a hard-label attack, which starts from an adversarial point and iteratively reduces the distance to the natural image. \cite{Carlini2017:bypass-detection} demonstrates that methods detecting adversarial examples can be defeated as well. 

\subsection{Adversarial Defenses}
Adversarial training \cite{intriguing_properties_dnns,explaining_harnessing_adv_egs} is one of the first defenses proposed and still remains one of the most effective defenses to this day. Madry \etal \cite{toward_robust_dnn} show that if the adversary is able to effectively solve the inner maximization problem, the DNN can adjust its parameters to withstand worst-case perturbations. Extensions of adversarial training have been proposed \cite{tramr2017ensemble,uesato2019labels,Shafahi2019:AT-for-free}. \cite{stutz2019confidencecalibrated} builds robustness by applying label smoothing to $\ell_{\infty}$ adversarial training. \cite{lamb2019interpolated} helps to mitigate adversarial training's negative effect on standard accuracy \cite{robustness_vs_accuracy}. Others \cite{gong2017adversarial,grosse2017statistical,metzen2017detecting,roth2019odds} attempt to detect adversarial examples before feeding them to the network. 

%In practice, adversarial training has been shown to be the best defense in many cases. Unfortunately, adversarial training is computationally costly. For each backwards pass through the network when training in the standard, supervised fashion, a network trained with adversarial training must perform many additional backwards passes. Practically, this means that adversarial training cannot be used in some real world scenarios.

As adversaries often exploit noisy and extreme gradients \cite{grad_mag_and_robustness}, a class of techniques has emerged that regularizes the gradients of the network in order to gain robustness. Ross \etal \cite{input_grad_reg} propose a variation of double backpropagation \cite{double_bp}, and show that regularizing the loss gradient is an effective defense against FGSM, JSMA, and the targeted gradient sign method. Similarly, Jakubovitz \etal \cite{jacobian_reg} propose to regularize the input-logits Jacobian matrix. Furthermore, Parseval Networks \cite{parseval_nets} constrain the Lipschitz constant of each layer. Cross-Lipschitz regularization \cite{cross_lipschitz} forces the differences between gradients of each class score function to be small.

\subsection{Relationship Between Adversarial Robustness and Interpretability} \label{subsec:adv_robustness_and_interp}
% It has been observed that when implementing these defenses, standard accuracy drops. In fact, Tsipras et al. find that robustness \textit{must} come at the cost of some generalization \cite{robustness_vs_accuracy}.

% They argue that undefended networks inch their accuracy a bit higher by learning to rely on features that are not robust to adversarial perturbations. In this same vein, Ilyas et al. \cite{ilyas2019adversarial} provide a new perspective on adversarial examples that explains their transferability.

% {\cite{ilyas2019adversarial} suggests that the brittleness results from the network learning to use non-robust features in the data. However, the reason for the prevalence of adversarial brittleness and how the human visual system (mostly) avoids this vulnerability remain open questions. }

Recently, it has been observed that robust networks tend to be more interpretable. Anil \etal \cite{lipschitz_nets_interp} remark that networks trained with Lipschitz constraints have gradients that appear more interpretable. Similarly, Ross \etal \cite{input_grad_reg} find that gradient regularized networks have qualitatively more interpretable gradient maps. Others \cite{robustness_vs_accuracy,prasad} note that the simple gradient salience maps generated from adversarially trained models are more interpretable than those generated from non-robust models. Tsipras \etal \cite{robustness_vs_accuracy} provide the hypothesis that models that can withstand adversarial examples have necessarily learned to rely on features invariant to adversarial perturbations. Because humans are naturally invariant to these perturbations, robust models tend to function more similarly to the human vision system than non-robust models. Etmann \etal \cite{robustness_interpretability} provide theoretical justification that robust linear models tend to have gradients that are co-linear with the input images, which is related to interpretability.

A couple of works explored the relationship between model robustness and interpretability from different perspectives than our own. \cite{Dong2017} explain the role of individual neurons with adversarial examples.  \cite{frgagile_interpretations} show that interpretations of neural networks are not immune to adversarial attacks. 

Using an approach similar to Generative Adversarial Networks, Chan \etal \cite{JARN} force the Jacobian of the network to contain information needed to reconstruct a natural image. The resulting network becomes robust to certain $\ell_\infty$ PGD attacks, especially when some adversarial training has been added. However, it remains obscure if the complex training procedure yields interpretable networks. A parallel work \cite{lanfredi2020quantifying} defines interpretability in binary classification as the alignment between the gradient direction and the line connecting a data point and its closest neighbor of the opposite class, and shows that improving this interpretability improves robustness over standard training. To the best of our knowledge, no prior work has demonstrated that forcing a model's gradients to be interpretable improves the model's robustness in multi-class classification. 

% \subsection{Gradient-based Regularization} \label{subsec:gradient_targets}
% Some works have trained networks to have gradients that match certain targets. A couple of works \cite{sobolev,attribution_priors} show that training networks with target gradients in addition to target labels results in improved generalization. A recent paper \cite{JARN} has shown that forcing the gradient of a network's loss function to contain all of the information required to reconstruct the input results in improved robustness. However, 

\section{Approach} \label{sec:approach}
The objective of our experiments is to determine if it is possible to make a model robust to adversarial perturbations by optimizing the model to have interpretable gradients. To this end, we supplement the standard cross-entropy loss function with two regularization terms that together encourage the simple gradient salience map for each data point to agree with an interpretable target interpretation.

We introduce the following notations. A data point, drawn from the data distribution $D$, consists of an input $\bm x \in \mathbb{R}^D$ and a label $\bm y \in \mathbb{R}^K$. Here $\bm y$ is a $K$-dimensional one-hot vector that contains a single 1 at the correct class $y_c$ and zeros at the other $K-1$ positions. The neural network $f_{\bm \theta}(\cdot)$ has parameters $\bm \theta \in \mathbb{R}^M$ and outputs the logits before the final softmax operation, so that the model prediction for $\bm{x}$ can be written as $\hat{\bm y} = \text{softmax}(f_{\bm \theta}(\bm x))$.
Additionally, the input-logits Jacobian matrix $\mathcal{J}(\bm x) \in \mathbb{R}^{K\times D}$ is computed as $\mathcal{J}(\bm x) = \partial f_{\bm \theta}(\bm x) / \partial \bm x$.\footnote{ $\mathcal{J}(\bm x)$ is distinct from the input-output Jacobian matrix $\mat{J}(\bm x)$. They are related by $\mat{J}(\bm x) = (\text{diag}({\hat{\bm y}})-{\hat{\bm y}}{\hat{\bm y}}^\top)\mathcal{J}(\bm x)$.} Of particular interest is the slice of the Jacobian matrix corresponding to the correct label, $\mathcal{J}_{c}(\bm x) = \langle \partial f_{\bm \theta}(\bm x)/ \partial \bm x\rangle_{[c,:]} $, also known as the simple gradient salience map \cite{simple_gradient}.

With standard supervised training, the optimal parameters $\bm \theta^*$ are found by minimizing the cross-entropy loss. 
\begin{equation} \label{eq:XE-loss}
\mathcal{L}_{XE}(\bm x, \bm y, \bm \theta) =  \sum_{i} y_i \log \hat{y}_i 
= y_{c} \log \hat{y}_{c}
\end{equation}
Assuming the availability of target interpretations $I(\bm x)$ for each $\bm{x}$ (covered in Section \ref{subsec:target_interpretation}), we can add two regularization terms to the standard loss in order to (1) encourage the network to align its gradients with the target interpretations and (2) restrict the magnitude of its simple gradient salience maps.
\begin{equation} \label{eq:proposed-loss}
    \mathcal{L}(\bm{\theta}) = \underset{(\bm{x},\bm{y}) \sim \mathbf{D}} {\mathbb{E}} 
\left[  
\mathcal{L}_{XE}(\bm{x},\bm{y}, \bm{\theta}) - 
\lambda_d \cos(\mathcal{J}_c (\bm x), I(\bm x)) + 
\lambda_m \| \mathcal{J}_c (\bm x) \|_F \right]
\end{equation}
In the above, $\cos(\mathcal{J}_c (\bm x), I(\bm x))$ is the cosine of the angle between the vectorized target interpretation $I(\bm{x})$ and the vectorized simple gradient $\mathcal{J}_{c}(\bm{x})$. The
coefficients $\lambda_d$ and $\lambda_m$ control the regularization strengths.
Given interpretable target interpretations, these terms encourage the gradients of the model to be interpretable. %Intuitively, the first regularization term forces the simple gradient salience maps and the target interpretations to have similar directions, and the second regularization term encourages the simple gradient salience maps to have small magnitudes.  %In this way, we can optimize the model being trained for interpretability and determine the degree to which model interpretability helps in terms of robustness to adversarial perturbations.

\subsection{Generating Target Interpretations} \label{subsec:target_interpretation}
Using the SmoothGrad method \cite{smoothgrad}, we extract target salience maps for each data point from a pretrained neural network (details in Section \ref{subsec:adversarial training}).
We choose SmoothGrad over other interpretation methods for two reasons. First, it satisfies the basic sensitivity and invariance properties \cite{sanity_checks_salience_maps,unreliability} discussed in Section~\ref{subsec:interpreting_dnn}, which assert that the interpretation is properly sensitive to the model and data distributions. Second, SmoothGrad can be understood as a method for canceling out the influence of small perturbations on the interpretation, which has the effect of drawing the interpretation closer to what humans find meaningful \cite{smoothgrad}.

The SmoothGrad method first samples $N$ points around a given input $\bm x$ from the standard Gaussian distribution and 
takes the mean of the simple gradient salience maps generated for each sample. Formally, having drawn $N$ independent $e_i \sim \mathcal{N}(0, \sigma^2)$, the interpretation $I^\text{SmG}(\bm x)$ is computed as the Monte Carlo expectation.
\begin{equation}
    I^\text{SmG}(\bm{x}) = \frac{1}{N}\sum_{i = 1}^N \mathcal{J}_{c}(\bm x + e_i)
    \label{eq:our-smoothgrad}
\end{equation}

In order to filter out small values that are usually ignored by a human observer, we further threshold the target salience map with its standard deviation. For each interpretation $I^\text{SmG}(\bm x)$, we compute the pixel-level standard deviation $\sigma_\text{S}$ and mean $\mu_\text{S}$. Any value in $I^\text{SmG}(\bm x)$ falling within the range $[\mu_\text{S} - \phi \sigma_\text{S}, \mu_\text{S} + \phi \sigma_\text{S}]$ is set to zero.  $\phi$ is a hyperparameter that determines the filtering strength. The thresholding can also be understood as a sparsification operation, like that used in $\ell_0$-regularized regression (e.g., \cite{Jain2016:hard-thresholding}), so that we can focus produced salience maps on the most important features of the images. 

\begin{figure*}
\setlength{\belowcaptionskip}{-10pt}
\centering
\includegraphics[width=1\linewidth]{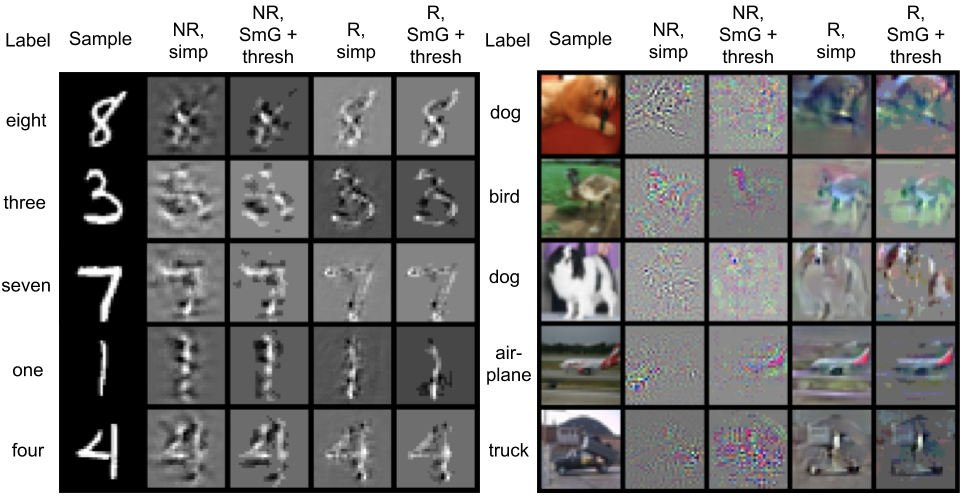}
\caption{MNIST (left half) and CIFAR-10 (right half) salience maps from adversarially trained networks (R) and networks with standard training (NR). ``simp'' and ``SmG'' denote target interpretations generated using simple gradient and SmoothGrad, respectively. The threshold $\phi$ is set to 1.}\label{fig:saliency_maps}
\end{figure*}

Figure \ref{fig:saliency_maps} contains examples of generated target interpretations. Observing the figure, we note that,  just as we expect,  the robust network produces more intepretable salience maps than the non-robust networks, especially for the CIFAR-10 images. However, even with the non-robust network, the salience maps from SmoothGrad + thresholding can sometimes identify important features, such as the head and neck of the bird (second image on the right) or the rear of the airplane (fourth image on the right). For MNIST images, negative values in the salience maps are shown as black spots, which indicate missing strokes that could change the classification. For example, if we fill in the black spots on the image of 3, we would transform it into 6. Overall,  the combined use of SmoothGrad and thresholding has been shown to erase the noisy components but retain the important parts of the interpretations.

\subsection{Adversarial Training} \label{subsec:adversarial training}
We create robust neural networks using adversarial training, one of the earliest and still most reliable defenses. The purpose of this model is to supply the target interpretations and serve as the upper bound for robustness in the experiments. 

We adopt a projected gradient descent (PGD) adversary that iteratively adds perturbations to an input sample to fool a model.
Formally, we let $\bm x_t$ denote the input after $t$ iterations of transformation and $\bm x_0 = \bm x$. After each perturbation is added, the data point is projected to the nearest point within an $\ell_2$ hypersphere with the radius $\epsilon$ around $\bm x_0$. This operation is denoted by the function $\text{clip}_{\bm x_0, \epsilon}(\cdot)$. The $\text{sign}(\bm z)$ function maps the $D$-dimensional vector $\bm z$ to the element-wise sign vector $\{-1, 1\}^D$. The iterative optimization can be characterized as 
\begin{equation}
    \bm x_{t+1} = \text{clip}_{\bm x_0, \epsilon}(\bm x_ {t} + \alpha \;  \text{sign}(\mathcal{L}_{XE}(\bm x_t, \bm y, \bm \theta))),
\end{equation}
\begin{equation}
\text{clip}_{\bm x_0, \epsilon}(\bm x^\prime) = 
\begin{cases} 
\bm x_0 + \frac{\bm x^\prime - \bm x_0}{\| \bm x^\prime - \bm x_0 \|_2} \epsilon & \text{if } \| \bm x^\prime - \bm x_0 \|_2 > \epsilon\\
\bm x^\prime & \text{otherwise.}
\end{cases}
\end{equation}
where $\alpha$ is the step size. After the adversarial samples are created at each iteration of training, they are used in place of the original samples.
% The PGD adversary began the process of finding an adversarial example by randomly perturbing an example. 
% We chose an $\ell_2$-bounded PGD adversary to train against following Tsipras et al. \cite{robustness_vs_accuracy}.
%showed that the interpretations generated from networks trained against this adversary produced salience maps that were more interpretable than salience maps generated from networks trained against other adversaries bounded by different distance metrics.

% As mentioned above, the perturbations existed in the $\ell_{2}$ norm with epsilon at 1.5 for MNIST and epsilon at .314 for CIFAR-10. The number of iterations for the attack was set to 40 for both. We chose this number of iterations and these epsilon values as Tsipras et al. demonstrated that these values provided a good balance between too much / too little of a change to the original samples \cite{robustness_vs_accuracy}. 
% It is worth noting that when training the network to be used to generate the target interpretations, the imperceptibility of the examples trained against is not important. In fact, it is reasonable to assume that training the network to be robust to large, noticeable perturbations will force the network to rely only on those features that are most invariant to the adversary's perturbations \cite{robustness_vs_accuracy, ilyas2019adversarial}.

\subsection{Jacobian Regularization}
Jacobian regularization \cite{jacobian_reg} is another defense technique, which supplements the original cross entropy loss with a regularization term. 
\begin{equation} \label{eq:jacobian-regularization}
    \mathcal{L}(\bm{\theta}) = \underset{(\bm{x},\bm{y}) \sim \mathbf{D}} {\mathbb{E}} 
\left[  
\mathcal{L}_{XE}(\bm{x},\bm{y}, \bm{\theta}) + \lambda_J \| \mathcal{J}(\bm x) \|_F \right]
\end{equation}
Where $\| \cdot \|_F$ is the Frobenius norm and $\lambda_J$ is a hyperparameter determining the strength of the regularization. Contrasting Eq. \ref{eq:jacobian-regularization} with Eq. \ref{eq:proposed-loss}, the regularization term in Jacobian Regularization suppresses all entries in the Jacobian matrix, whereas Interpretation Regularization is only concerned with the one slice of the Jacobian that corresponds to the correct label. 

Hoffman \etal \cite{jacobian_reg_approx} demonstrate that the Frobenius norm of the Jacobian can be approximated using random projections.
\begin{equation}
    \| \mathcal{J}(\bm x) \|_F \approx \sqrt{ \frac{1}{n_{proj}}\sum_{\mu=1}^{n_{proj}}\left[ \frac{\partial(\bm{v}^\mu \cdot f_{\bm \theta}(\bm x))}{\partial x} \right]^2}
\end{equation}
Here the random projection vector $\bm{v}^\mu$ is drawn from the $(K-1)$-dimensional unit sphere for every training iteration. In practice, Hoffman \etal shows that even a single random projection vector effectively suppresses the Jacobian norm.

\section{Experiments} \label{sec:experiments}
In this section, we present three experiments. In the first two experiments, we compare Interpretation Regularization with adversarial training, Jacobian regularization, and ablated variants on MNIST and CIFAR-10. After that, we further explore the role of the target interpretation by using target interpretations permuted to different degrees. 
Finally, we discuss the results and their implications. 

% first discuss the setup for the experiments --- with the MNIST and CIFAR-10 datasets --- in which we train networks to match various sets of target interpretations. Then we discuss the setup for a third experiment involving MNIST in which we train a network to match sets of target interpretations that have been permuted to different degrees. After this

\subsection{General Setup}
We first describe some general setup that applies to all experiments. The pixel values for each image were normalized to the range $[0,1]$. We used $\ell_{2}$ adversarial training (AT) throughout, as preliminary results showed Interpretation Regularization worked better with interpretation targets from $\ell_{2}$ AT than $\ell_{\infty}$ AT. Note that, on MNIST, $\ell_2$ projected gradient descent (PGD) adversaries tend to be less effective than $\ell_{\infty}$ adversaries and provide weaker defenses \cite{toward_robust_dnn,l2mnist}. We attacked all networks using both $\ell_{2}$ and $\ell_{\infty}$ attacks. 

The target interpretations were generated in the following manner. We followed the original recommendation for SmoothGrad \cite{smoothgrad} and set the noise level at $\sigma$ = 0.15 and the number of samples $N$ to 50. The filtering threshold was set to $\phi=1$ (See Section \ref{subsec:target_interpretation} for details). We extracted simple gradient and SmoothGrad saliency maps from $\ell_2$ adversarially trained networks as well as non-robust networks that were trained only on natural images. In addition, we also created a \textit{complete} random permutation (permutation probability at 1.0) of the robust SmoothGrad interpretations. 

For each dataset, we created baselines using adversarial training and  Jacobian regularization. In order to facilitate meaningful comparisons, all networks were trained to have roughly the same validation loss on natural images. 

Due to space limitations, we use some shorthands in tables and figures to denote these configurations. ``R'' and ``NR'' denote target interpretations generated from robust and non-robust (standard trained) networks, respectively. Simple gradient and SmoothGrad are denoted by ``simp'' and ``SmG'', respectively. ``perm'' means that each target interpretation was completely randomly permuted. ``IR'' indicates Interpretation Regularization, ``AT'' indicates adversarial training, and ``JR'' stands for Jacobian regularization \cite{jacobian_reg_approx}.

\subsection{MNIST} \label{subsec:mnist_experiments}

The MNIST experiments were set up in the following ways. The convolutional neural network (CNN), taken from \cite{simple_cnn}, had two convolutional layers of 32 and 64 filters of size $5\times5$ and stride 1. Each convolutional layer was followed by max-pooling with a $2\times2$ kernel and a stride of 2. The last pooling layer fed into a dense layer with 1024 neurons. Dropout with $p=0.5$ was applied before the final dense layer of 10 neurons and the softmax operation. All layers except the last employed the ReLU activation function. All experiments used SGD with momentum at 0.9, an initial learning rate of 0.01 that decayed to zero using the cosine schedule, a batch size of 50, and a maximum number of epochs of 100.

Following \cite{robustness_vs_accuracy}, we extracted robust interpretations from an adversarially trained CNN with a randomly initialized PGD adversary using an $\ell_2$ radius of 1.5 and 40 iterations of PGD. Tspiras \etal \cite{robustness_vs_accuracy} qualitatively showed that training against this adversary produced networks that had interpretable simple gradient salience maps. This network, along with a second network trained with a PGD adversary with an $\ell_2$ radius of 2.5 and 40 iterations of PGD, serve as baselines.

% It is worth noting that when training the network to be used to generate the target interpretations, the imperceptibility of the examples trained against is not important. In fact, it is reasonable to assume that training the network to be robust to large, noticeable perturbations will force the network to rely only on those features that are most invariant to the adversary's perturbations \cite{robustness_vs_accuracy, ilyas2019adversarial}.
%After creating target salience maps for each input sample in the MNIST training set using the robust network and SmoothGrad, a new network with the same architecture was trained using the modified loss described in equation \eqref{eq:1}, with $I(\cdot)$ being the simple gradient method. 

Using the robust network's simple gradient maps as the target interpretations (R, simp), we performed a grid search to find the best combination of $\lambda_d$ (which controls the strength of the gradient alignment with $I(x)$) and $\lambda_m$ (which controls the magnitude of $\| \mathcal{J}_{c(\bm x)}(\bm x) \|_F$). $\lambda_d$ at 3.0 and $\lambda_m$ at 0.15 produced the best results. For the experiments with the other four sets of target interpretations, we simply reused these hyperparameter values and did not do any additional tuning. 
For baselines, we created adversarially trained networks with radii of $1.5$ and $2.5$. For Jacobian regularization, we performed a search across $\lambda_J$ and found 0.32 to produce the best results.

% In order to investigate the effects of different defenses and regularization methods, a number of baselines were produced, including standard training with only natural images (Standard Training), adversarial training with the PGD adversary described above (AT PGD40, $\ell_2$, $\epsilon=2.5$), and Jacobian Regularization (JR). 

% In addition to training the simple CNN with the Interpretation Regularization loss, we trained separate copies of the architecture with only natural images, with PGD-perturbed training data (using the PGD adversary described above), and with Jacobian regularization. We attempted to choose hyperparameters for Jacobian and Interpretation Regularization such that the training losses were similar. Our hope was that in doing this, the standard test accuracies would be roughly equivalent and the adversarial accuracies could be compared in a meaningful way.

% In order to better understand why Interpretation Regularization was working, we took the target interpretations and randomly permuted their values. We then trained a network to match these permuted target interpretations.

\subsection{CIFAR-10} \label{subsec:cifar10_experiments}
For all experiments with the CIFAR-10 dataset, the Wide ResNet (WRN) 28$\times$10 architecture \cite{wide_resnet}, a large network with 36.5 million parameters, was used. We adopted weight decay of $5\mathrm{e}{-4}$, dropout rate of 0.3, SGD with Nesterov momentum at 0.9, an initial learning rate of 0.1 decaying to zero under the cosine schedule, batch size of 128, and 200 training epochs. 
During training, we augmented the training data with random cropping and random horizontal flips, and with IR, each target interpretation was transformed in the same way as its corresponding input image.

We obtained robust networks using adversarial training with PGD, two different $\ell_2$ radii of $80/255$ and $320/255$, and 7 PGD iterations. Again, this is the same setup which showed qualitatively interpretable gradients in  \cite{robustness_vs_accuracy}. For Jacobian regularization, we adopted the approximation from \cite{jacobian_reg_approx} with $n_{proj}=1$, and we did a search across $\lambda_J$, finding $\lambda_J = 0.1$ and $0.03$ to produce the best results. 
For the simple gradient map of the robust network (R, simp), $\lambda_d$ at 0.75 and $\lambda_m$ at 0.005 or 0.02 were found by a grid search to produce good results and validation losses on natural images that were comparable with the two adversarially trained networks. These hyperparameters were used across the other four Interpretation Regularization experiments with no further tuning. 

% To provide baselines for comparison, other copies of the architecture were trained using standard training, adversarial training (two networks: one with an adversary with a smaller $\epsilon=80/255$ and a second network with an adversary with a larger $\epsilon=320/255$), and Jacobian regularization with the approximation from \cite{jacobian_reg_approx} with $n_{proj}=1$. As can be seen in Table \ref{tab:results-cifar}, two strata of accuracies are considered. Again, in order to facilitate meaningful comparisons, we trained all networks to have roughly the same validation loss on natural images. 
% Again, we tuned the hyperparameter $\lambda_J$ for Jacobian regularization so the final training loss matched those networks trained with Interpretation Regularization to facilitate meaningful comparison.

% \subsection{Generating Target Interpretations From a Non-robust Network}
% \hl{have a subsection to enumerate the baselines} When the target interpretations are generated using a network that achieves high test set accuracy but low adversarial accuracy, the model that is regularized so that its interpretations match the target interpretations is \textit{not} robust. This is when there is no thresholding on the target interpretations.

% \subsection{Randomly Permuting the Values in the Target Interpretations}

\subsection{MNIST with Permuted Target Interpretations} \label{subsec:variable_permute_setup}

\begin{figure}
    \setlength{\belowcaptionskip}{-5pt}
    \centering
    \includegraphics[width=.99\textwidth]{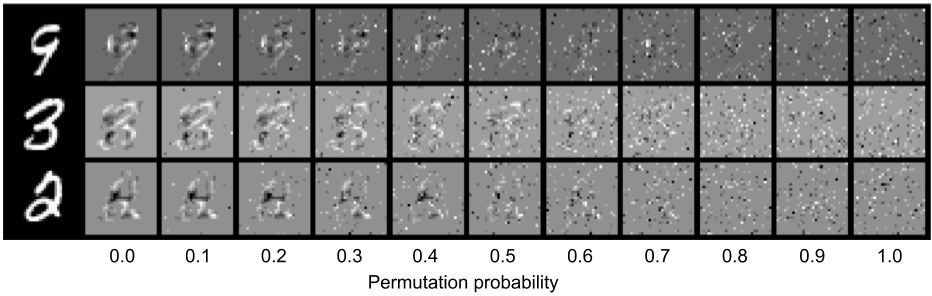}
    % \vspace{1cm}
    \caption{Three SmoothGrad target interpretations with varying degrees of permutation. The original images (with labels of 9, 3, and 2, from top to bottom) are in the leftmost column.
    }
    \label{fig:variable_permute}
\end{figure}

To further investigate the effects of the target interpretation, we conducted an additional experiment with new sets of permuted target interpretations. We first extract the SmoothGrad interpretations from the robust network, which was adversarially trained on the MNIST dataset as in Section \ref{subsec:mnist_experiments}. After that, we randomly permuted from 10\% to 100\% of pixels in each  interpretation to obtain ten new sets of interpretations. Each of these ten sets was then used as target interpretations for Interpretation Regularization. Figure \ref{fig:variable_permute}   shows some example target interpretations from each of the ten sets. 
In this way, the mean and standard deviations of the pixel values in each target interpretation was held constant, but the semantic patterns in the target interpretations were disrupted to varying extents. 
%After that we use Intepretation Regularization to train the network to match the permuted interpretation and test their robustness under a \hl{what kind} adversary. 

% For fair comparisons, we train models with differently permuted target interpretations to comparable training loss. 
The optimal $\lambda_d$ and $\lambda_m$ from Section \ref{subsec:mnist_experiments} were used without any changes. In other words, the only hyperparameter that changed across networks being trained in this experiment was the permutation probability for the target interpretations. The results for this experiment can be found in Figure \ref{fig:permute_graph}.

\subsection{Results} \label{subsec:results}

\begin{table*}[!t]\centering
\small
\begin{tabular}{@{}p{2.85cm}P{1.5cm}P{1.2cm}P{0.8cm}P{0.8cm}P{0.8cm}P{0.8cm}@{}}\toprule
\multirow[c]{3}{1.5cm}[-0.15in]{\textbf{Training Technique}} & \multirow[c]{3}{1.5cm}[-0.15in]{\textbf{Standard Accuracy}} &\multicolumn{5}{c}{\textbf{Adversarial Accuracy}} \\ \cmidrule{3-7}
& & PGD40, $\ell_2$ norm & \multicolumn{4}{c}{PGD40, $\ell_\infty$ norm} \\
\cmidrule(r){3-3} \cmidrule(l){4-7}
& & $\epsilon=1.50$ & $0.10$ & $0.20$ & $0.25$ & $0.30$\\
\midrule
Standard Training &             99.50 & 78.41 & 78.91 & 9.32 & 3.57 & 1.71\\
AT (PGD40, $\ell_2$, $\varepsilon$=1.5) & 99.39 & \textbf{89.88} & \textbf{96.60} & \underline{73.01} & 32.07 & 5.46 \\
AT (PGD40, $\ell_2$, $\varepsilon$=2.5) & 98.29 & \underline{88.06} & \underline{94.23} & \textbf{76.39} & \textbf{49.98} & \underline{10.94} \\
% JR & 99.37 & 83.25 &  93.94 &  52.26 &  15.33 &  1.50 \\
% JR & 99.08 & 84.00 &  93.50 &  59.35 &  25.92 &  5.03 \\
JR &                                    98.12 & 82.64 &  91.15 &  60.58 &  29.41 &  6.54 \\
% IR (R, simp) & 98.35 & 83.18 & 91.91 &  63.83 &  38.53 &  12.23 \\
IR (R, perm, SmG) &             97.28 & 78.75 &  88.11 &  54.36 &  26.26 &  7.58 \\
IR (NR, simp) &                 98.05 & 79.57 & 87.82 & 50.20 & 23.71 &  2.29 \\
IR (NR, SmG) &                  98.04 & 81.22 & 90.39 &  55.98 &  28.70 &  4.98 \\
IR (R, simp) &                  98.12 & 84.24 &  91.60 &  64.03 &  \hspace*{-.125cm}*37.90 &  \hspace*{-.125cm}*10.86 \\
IR (R, SmG) &                   98.18 & \hspace*{-.125cm}*85.25 &  \hspace*{-.125cm}*92.35 &  \hspace*{-.125cm}*66.92 &  \underline{41.22} &  \textbf{11.52} \\
\bottomrule
\end{tabular}
\caption{Mean MNIST adversarial accuracies on the test set averaged over 3 random restarts of the attacks. The first, second, and third highest accuracies for each adversary are bolded, underlined, and asterisked, respectively.}
\label{tab:results-mnist}
\end{table*}

\begin{table*}[!t]\centering
\small
\begin{tabular}{@{}p{3.15cm}P{1.4cm}P{1.4cm}P{1.2cm}P{1.2cm}P{1.2cm}@{}}\toprule
\multirow{3}{3.5cm}[-0.05in]{\textbf{Training Technique}} &\multirow{3}{1.7cm}[-0.05in]{\textbf{Standard Accuracy}} &\multicolumn{4}{c}{\textbf{Adversarial Accuracy}} \\\cmidrule{3-6}
& &\multicolumn{2}{c}{PGD40, $\ell_2$ norm} &\multicolumn{2}{c}{PGD40, $\ell_\infty$ norm} \\\cmidrule(r){3-4} \cmidrule(l){5-6}
& &$\epsilon$ = 80/255 &$320/255$ & $4/255$ &$8/255$ \\\midrule
Standard Training & 95.00 & 11.84 & 0.00 & 0.74 & 0.01 \\\midrule
AT (PGD7, $\ell_2$, $\varepsilon$=320/255) & 76.04 & \textbf{67.57} & \textbf{58.29} & \textbf{59.05} & \textbf{38.4} \\
JR & 78.93 & 61.41 & 8.35 & 42.22 & 11.23 \\
IR (R, simp) & 78.68 & \underline{63.74} & \underline{12.56} & 47.18 & 16.28 \\
IR (R, SmG) & 78.97 & 62.64 & 12.42 & \underline{48.17} & \underline{18.78} \\\midrule
AT (PGD7, $\ell_2$, $\varepsilon$=80/255) & 90.34 & \textbf{75.14} & \textbf{20.44} & \textbf{59.43} & \textbf{25.37} \\
JR & 85.41 & 58.96 & 3.32 & 32.27 & 2.51 \\
IR (NR, simp) & 81.43 & 45.04 & 0.18 & 16.69 &  0.40 \\
IR (NR, SmG) & 84.39 & 53.84 & 0.56 & 28.00 & 1.84 \\
IR (R, perm, SmG) & 85.70 & 58.20 & 1.91 & 29.19 & 1.87 \\
IR (R, simp) & 85.39 & 62.45 & 5.35 & 39.93 & 8.2 \\
IR (R, SmG) & 85.69 & \underline{63.71} & \underline{7.98} & \underline{46.64} & \underline{14.25} \\

% IR (NR, SmG) \\
\bottomrule
\end{tabular}
\caption{Mean CIFAR-10 adversarial accuracies on test set averaged over 3 random restarts of the attacks. Within each group of models, the highest and second highest accuracies for each adversary are bolded and underlined, respectively.}
\label{tab:results-cifar}
\end{table*}

\begin{figure}
    \setlength{\belowcaptionskip}{-10pt}
    \centering
    \includegraphics[width=.99\textwidth]{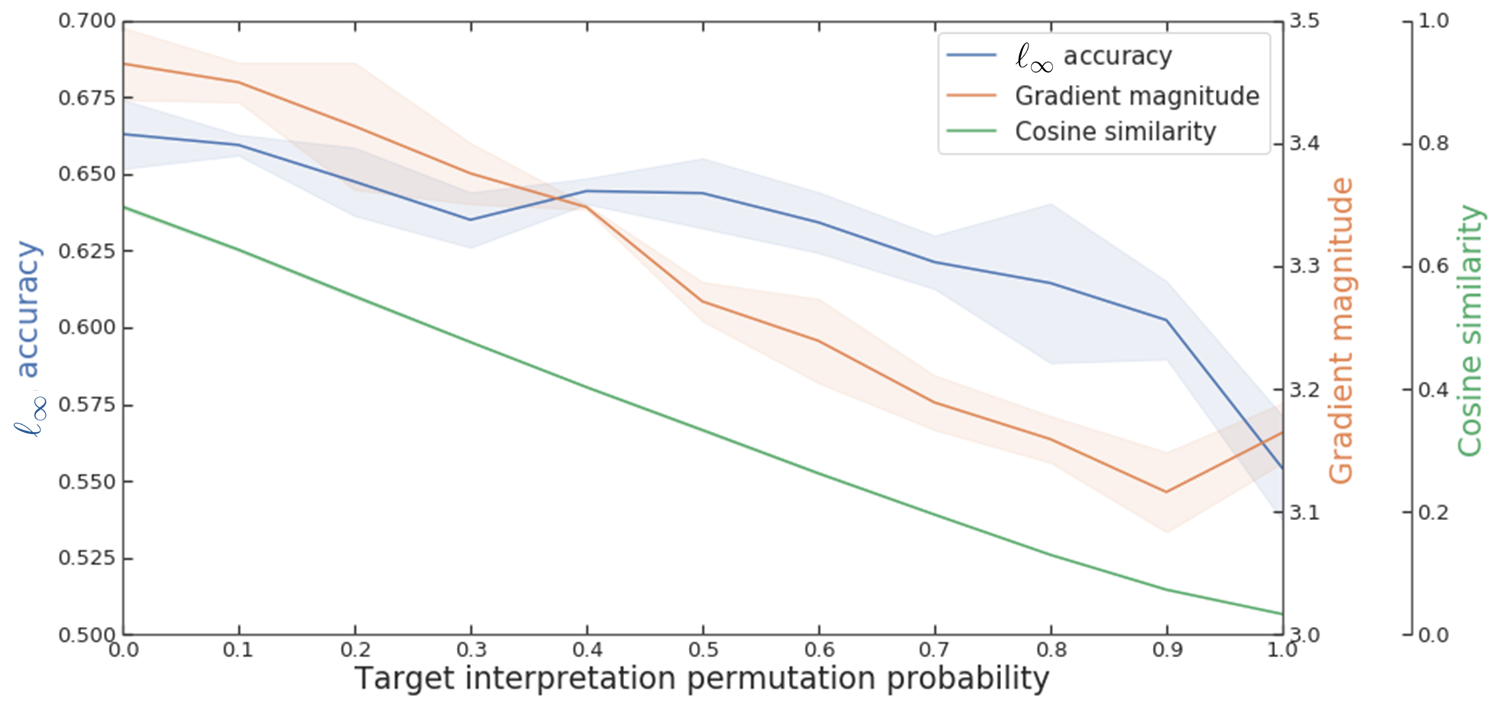}
    % \vspace{1cm}
    \caption{Adversarial accuracy under differently permuted target interpretations. On the y-axis: PGD-40 $\ell_{\infty}$ ($\varepsilon = .2$) adversarial accuracy, gradient magnitudes $\gradient$, and average cosine similarity between target interpretations and the gradients. The x-axis represents the permutation probability for the target interpretations. 95\% confidence interval shading over $n=3$ independently trained networks.
    }
    \label{fig:permute_graph}
\end{figure}

\begin{figure}
\centering
\begin{subfigure}{.5\textwidth}
    \centering
    \includegraphics[width=1\textwidth]{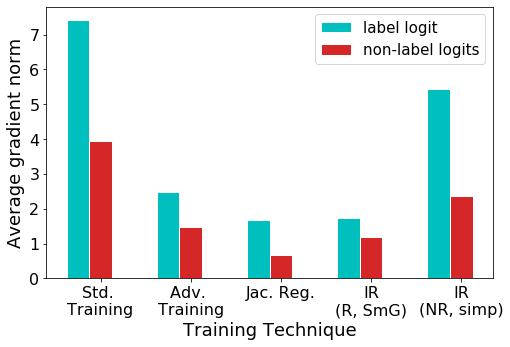}
\end{subfigure}%
\begin{subfigure}{.5\textwidth}
    \centering
    \includegraphics[width=1\textwidth]{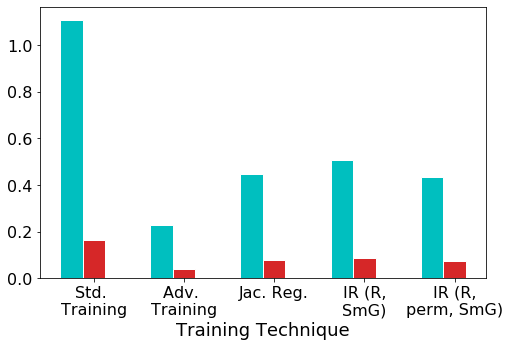}
    % \caption{Mean gradient norms of the label logit, $\| \nabla f_{c(\bm{x})}(\bm{x}) \|_2$, and non-label logits, $\| \nabla f_{\bm y\setminus c(\bm{x})}(\bm{x}) \|_2$, for CIFAR-10 test samples.}
\end{subfigure}
\caption{Mean Frobenius norms of the Jacobians for the label logit, $\| \mathcal{J}_{c}(\bm x) \|_F$, and non-label logits $\| \mathcal{J}_{\neq c}(\bm x) \|_F$ for MNIST (left) and CIFAR-10 (right).}
\label{fig:gradient-analysis}
\end{figure}

Tables \ref{tab:results-mnist} and \ref{tab:results-cifar} report the standard and adversarial accuracies under different $\ell_2$ and $\ell_{\infty}$ norm constraints on MNIST and CIFAR-10. Since all adversarial training was performed with an $\ell_2$ adversary, and the IR-trained models used the target interpretations extracted from the $\ell_2$-trained models, the attacks from the $\ell_\infty$ adversaries create challenging cross-norm attack scenarios.  

With unperturbed data, standard training achieves the highest  accuracy and all defense techniques degrade the performance. The adversarial attacks prove effective, resulting in substantial performance degradation of the standard model. Three of the four attacks on CIFAR-10 brought the standard model's accuracy to below 1\%. For most attacks, adversarial training yields the highest adversarial performance and Interpretation Regularization is the second best. However, in the two highest difficulty settings on MNIST ($\ell_\infty$ norms $0.25$ and $0.30$), IR begins to surpass AT and becomes the most robust network under $0.30$. 
In addition, IR outperforms JR on all attacks. The performance differences between the two methods range from 1.2\% to 11.81\% on MNIST and from 2.34\% to 14.37\% on CIFAR-10. 

\sloppypar{The best IR performance is achieved, in almost all cases, by IR(R, SmG), which uses interpretations from SmoothGrad and the AT network. The performance of IR(R,simp) and IR(R, SmG) diverge the most, by 6.67\% and 6.05\%, when the interpretations are derived from $\ell_2$ AT with a low $\epsilon=80/255$ and attacked by the $\ell_\infty$ adversary. That is, when the adversarial training and attack have the most mismatch. On the other hand, IR(R,simp) has a slight edge of 1.1\% or less over IR(R, SmG) when the adversarial training uses $\ell_2$ adversaries with a large $\epsilon=320/255$ and the attacks come from $\ell_2$ adversaries. }
Finally, interpretations from non-robust models offer some robustness over standard training, but they compare unfavorably with JR or permuted interpretations. 

Figure \ref{fig:permute_graph} shows the effects of random permutation on the interpretation. The overall trend is quite clear: greater permutation causes lower adversarial accuracy. %Interestingly, this drop occurs despite the fact that the gradient magnitude drops as well, and it is well understood that gradient magnitude and robustness are inversely correlated \cite{jacobian_reg}.
Furthermore, as the permutation increases, the network becomes less and less able to align its gradients with the target interpretations. Not included in the graph are the standard accuracies of the networks; these accuracies trend monotonically downward as well, beginning at with an average of $98.13\%$ and ending with an average of $97.37\%$, and the Pearson correlation coefficient between permutation probability and standard accuracy is $-0.93$.

\subsection{Discussion} \label{subsec:discussion}

\textbf{Disentangling the effects of Jacobian norms and target interpretations.} \label{subsubsec:disentangling_mag_pattern}In the introduction, we showed that in order to defend against an arbitrary perturbation $\bm \delta$, we could suppress the Jacobian's singular values $\normtwo{\bm s}$, which is equivalent to  the Jacobian's Frobenius norm. This gives us Jacobian regularization. To verify that the suppression has happened, we plot the Frobenius norm of the input-logits Jacobians in Figure \ref{fig:gradient-analysis}. We separate the Jacobian slices that correspond to the correct class, $\mathcal{J}_c(\bm{x})$, and those of the incorrect classes $\mathcal{J}_{\neq c}(\bm{x})$. The results are averaged over all training samples. For the incorrect class slices, the results are also averaged over all output logits that differ from the ground truth. 

We find that almost all defense methods reduce the norms of Jacobians compared to standard training. Previously, we also observed that IR with permuted target interpretations can provide some adversarial robustness, even though it performs worse than JR. We attribute this effect to the fact that, even with a completely uninformative target interpretation, IR still decreases the Frobenius norm of Jacobians, which can improve robustness. 
In addition, the results show that for most models, the correct-class Jacobian norm was much larger than wrong-class Jacobian norms. This helps explains why IR is effective when it only constrains the correct-class Jacobian norm whereas JR constrains all slices of the Jacobian. 

However, it is also worth noting that lower norms do not always lead to better adversarial performance. In both MNIST and CIFAR-10 experiments, JR produces lower Jacobian norms than IR, but is consistently outperformed across all attacks. This indicates there are other factors at play. 

To further disentangle the effects of Jacobian norms and the interpretability of the Jacobians, we examine how degrees of random permutation affect adversarial robustness in IR (shown in Figure \ref{fig:permute_graph}). As the proportion of permuted pixels increases, the network gradually becomes less capable of withstanding $\ell_\infty$ attacks. Nevertheless, the reduction in robustness happens while the gradient magnitude (Jacobian's norm) decreases. This behavior cannot be explained from the perspective of Jacobian regularization or the minimization of $\normtwo{\bm s}$. With the other hyperparameters and training losses kept equal, we attribute the decrease in performance to the decline in quality of target interpretations.

\textbf{The quality of the interpretation matters.} \label{subsubsec:interp_quality}
We now examine how the interpretability of the Jacobian contributes to adversarial robustness. In the analysis in the introduction, we inferred that it is important to allocate the available budget of $\normtwo{\bm s}$ carefully in order to maximize predictive performance. Inspired by \cite{ilyas2019adversarial}, we conjecture that an interpretable Jacobian, which selects features that humans regard as important to the prediction, should provide adversarial robustness. 

Empirical evidences from the MNIST and CIFAR-10 experiments strongly corroborate this argument. First, models trained with target interpretations from robust models consistently outperform target interpretations from non-robust models. Second, random permutation of the interpretations causes significant performance drop. The final and the most compelling observation is that, in most cases, SmoothGrad interpretations perform better than simple gradient maps from both robust and non-robust models. This is especially pronounced when the attack uses a large $\ell_\infty$ perturbation, which delivers severe attacks for robust models trained with $\ell_2$ perturbations. In the MNIST experiments with $\ell_\infty$ radius of 0.25, IR(R,SmG) beats the AT network trained with $\ell_2$ radius of 1.5, from which the target interpretations for IR are extracted. Moreover, at $\ell_\infty$ radius of 0.30, IR(R,SmG) obtains the best robustness, surpassing even the AT network trained with $\ell_2$ radius of 2.5.  
With CIFAR-10, under $\ell_\infty$ perturbations, the performance of IR(R,SmG) always exceeds that of IR(R,simp). 

We ascribe the strength of SmoothGrad to the fact that it removes noise from the interpretation and creates more human-like interpretations than simple gradient. A qualitative observation of Figure \ref{fig:saliency_maps} suggests that SmoothGrad interpretations on MNIST are consistent with human intuition. For example, the black spots (negative gradient values) for the digit 3 indicate key differences between 3 and the digits 6 or 8 and thus supply important features for classification. Similarly, the black spots around the top of the digit 4 highlight the differences with the digit 9. 
The strong SmoothGrad performance shows that interpretability is directly correlated with adversarial robustness and Interpretation Regularization attains more than just the distillation of adversarially trained models.

\section{Conclusion} \label{sec:conclusion}
The abundance of adversarial attacks and the general difficulty of interpreting a DNN's predictions are two issues that render some applications of artificial intelligence impractical in the eye of the general public. The literature suggests that these two issues may be closely related, as works have indicated qualitatively that adversarial defenses techniques, such as adversarial training \cite{robustness_vs_accuracy}, Jacobian regularization  \cite{input_grad_reg}, and Lipschitz constraints \cite{robustness_interpretability} produce models that have salience maps that agree with human interpretations. 

These findings naturally lead one to wonder if the converse is true; if we force a neural network to have interpretable gradients, will it then become robust? We devise a technique called Interpretation Regularization, which regularizes the gradient of a model to match the target interpretation extracted from an adversarially trained robust model. The new model performs better than Jacobian regularization, which applies more constraints than Interpretation Regularization. Most importantly, applying the network interpretation technique SmoothGrad \cite{smoothgrad} improves robustness over the simple gradient interpretation technique, and in few cases, over the AT networks from which the target interpretations are extracted. These results suggest Interpretation Regularization accomplishes more than distilling existing robust models. 

In the discussion, we carefully disentangle two factors that contribute to the effectiveness of Interpretation Regularization: the suppression of the gradient and the selective use of features guided by high-quality interpretations. With the two factors, we manage to explain model behaviors under various settings of regularization and target interpretation. We believe this study provides useful insights into the research of adversarial defenses and interpretation methods. The joint investigation of these two issues will continue to foster our understanding of deep neural networks. 

\vspace{.2cm}
\noindent \textbf{Compliance with Ethical Standards} 

\textbf{Conflicts of interest}: The work done by Adam Noack was funded by the NSF Center for Big Learning (CBL) and a grant from the Air Force Research Laboratory and Defense Advanced Research Projects Agency, under agreement number FA8750-16-C-0166, subcontract K001892-00-S05. Isaac Ahern's work was funded by the NSF CBL. Dejing Dou was originally funded by the NSF CBL and now works at Baidu. Boyang Li originally worked with Baidu, but now works at Nanyang Technological University. 

\textbf{Ethical approval}: This article does not contain any studies with human participants or animals performed by any of the authors.

\vspace{.2cm}
\noindent \textbf{Funding} This work was funded by the NSF Center for Big Learning and a grant from the Air Force Research Laboratory and Defense Advanced Research Projects Agency, under agreement number FA8750-16-C-0166, subcontract K001892-00-S05.

% % For one-column wide figures use
% \begin{figure}
% % Use the relevant command to insert your figure file.
% % For example, with the graphicx package use
%   \includegraphics{example.eps}
% % figure caption is below the figure
% \caption{Please write your figure caption here}
% \label{fig:1}       % Give a unique label
% \end{figure}
% %
% % For two-column wide figures use
% \begin{figure*}
% % Use the relevant command to insert your figure file.
% % For example, with the graphicx package use
%   \includegraphics[width=0.75\textwidth]{example.eps}
% % figure caption is below the figure
% \caption{Please write your figure caption here}
% \label{fig:2}       % Give a unique label
% \end{figure*}
% %
% % For tables use
% \begin{table}
% % table caption is above the table
% \caption{Please write your table caption here}
% \label{tab:1}       % Give a unique label
% % For LaTeX tables use
% \begin{tabular}{lll}
% \hline\noalign{\smallskip}
% first & second & third  \\
% \noalign{\smallskip}\hline\noalign{\smallskip}
% number & number & number \\
% number & number & number \\
% \noalign{\smallskip}\hline
% \end{tabular}
% \end{table}

%\begin{acknowledgements}
%\end{acknowledgements}


\bibliographystyle{spmpsci}      % mathematics and physical sciences
\bibliography{biblio}   % name your BibTeX data base

% % Non-BibTeX users please use
% \begin{thebibliography}{}
% %
% % and use \bibitem to create references. Consult the Instructions
% % for authors for reference list style.
% % Format for Journal Reference
% Author, Article title, Journal, Volume, page numbers (year)
% Author, Book title, page numbers. Publisher, place (year)
% \end{thebibliography}

\end{document}